\pdfoutput=1

\documentclass[11pt]{article}

\usepackage[final]{acl}

\usepackage{times}
\usepackage{latexsym}

\usepackage[T1]{fontenc}

\usepackage[utf8]{inputenc}

\usepackage{microtype}

\usepackage{inconsolata}

\usepackage{graphicx}
\usepackage{dashrule} 
\usepackage{enumitem} 
\usepackage{booktabs} 
\usepackage{multirow}
\usepackage{multicol}
\usepackage{subfig}
\usepackage{amssymb}
\usepackage{amsmath}
\usepackage{xcolor}
\usepackage{longtable}
\usepackage[most]{tcolorbox}
\usepackage{pifont}

\title{Understanding the Modality Gap: An Empirical Study on the Speech-Text Alignment Mechanism of Large Speech Language Models}

\author{Bajian Xiang \quad Shuaijiang Zhao \quad Tingwei Guo \quad Wei Zou\thanks{Now at Bairong Inc., Beijing, China. Email: \texttt{wei.zou@brgroup.com}.} \\
Beike Inc., Beijing, China \\
\texttt{\{xiangbajian001, guotingwei002\}@ke.com}
}

\newcommand{\cmark}{\textcolor{green!60!black}{\ding{51}}}
\newcommand{\xmark}{\textcolor{red}{\ding{55}}}
\definecolor{beikeblue}{RGB}{23,147,230}

\begin{document}
\maketitle
\begin{abstract}
End-to-end Large Speech Language Models (LSLMs) have demonstrated impressive conversational generation abilities, yet consistently fall short of traditional pipeline systems on semantic understanding benchmarks. In this work, we reveal through systematic experimentation that although LSLMs lose some text input performance after speech-text alignment training, the performance gap between speech and text inputs is more pronounced, which we refer to as the \textbf{modality gap}. 
To understand this gap, we analyze both coarse- and fine-grained text and speech representations. 
At the coarse-grained level, representations of speech and text in deeper layers are found to be increasingly aligned in direction (cosine similarity), while concurrently diverging in magnitude (Euclidean distance). We further find that representation similarity is strongly correlated with the modality gap.
At the fine-grained level, a spontaneous token-level alignment pattern between text and speech representations is observed.
Based on this, we introduce the Alignment Path Score to quantify token-level alignment quality, which exhibits stronger correlation with the modality gap. 
Building on these insights, we design targeted interventions on critical tokens through angle projection and length normalization. These strategies demonstrate the potential to improve correctness for speech inputs. Our study provides the first systematic empirical analysis of the modality gap and alignment mechanisms in LSLMs, offering both theoretical and methodological guidance for future optimization.
\end{abstract}

\section{Introduction}

The emergence of Large Speech Language Models (LSLMs) has revolutionized human-computer interaction by enabling direct processing of both speech representations and text inputs, subsequently generating textual or spoken outputs \cite{bu2024roadmapsuperhumanspeechunderstanding-survey2}.
Compared to traditional pipeline architectures that sequentially chain Automatic Speech Recognition (ASR), Large Language Models (LLMs), and Text-To-Speech (TTS) components, end-to-end LSLMs offer significant advantages, including reduced latency, inherent error resilience, and more expressive speech synthesis capabilities \cite{ji2024wavchatsurveyspokendialogue-survey3}. 

Recent studies in LSLMs have focused on aligning speech modalities with text space through speech tokenizers \cite{zhang2023speechgpt} or encoder-based approaches \cite{fang2024llama, zhao2024advancing, Qwen-Audio}. These cross-modal alignment strategies aim to harness the linguistic capabilities of pretrained LLMs while integrating speech processing functionalities \cite{cui2025recentadvancesspeechlanguage-survey1}.

However, significant performance disparities persist between LSLMs and conventional pipeline models in semantic understanding tasks. Benchmark results from VoiceBench \cite{chen2024voicebench} reveal a striking contrast: the \textit{Whisper-large-v3 + LLaMA-3.1-8B} pipeline achieves 79.06 overall score, while its LSLM counterpart \textit{LLaMA-Omni} \cite{fang2024llama} scores merely 37.51. This pattern continues in Uro-bench \cite{yan2025urobench} evaluations, where the \textit{Whisper-large-v3 + Qwen2-7B-Instruct} pipeline attains 78.13 overall score compared to \textit{Freeze-Omni}'s \cite{wang2024freezeomnismartlowlatency} 48.28, despite both systems employing the same underlying LLM. Notably, while Uro-bench's dependence on transcribed speech outputs might inherently favor pipeline architectures, the magnitude of these performance drops remains substantial and warrants investigation.

Nevertheless, contemporary investigations into LSLMs remain largely confined to engineering practices, adopting unverified integrated solutions spanning training stages, dataset scales, parameter-efficient strategies, and multimodal objectives  without systematic analysis of their individual contributions or synergistic effects \cite{chu2024qwen2audio, zhong2024lyraefficientspeechcentricframework, liu2025olapushingfrontiersomnimodal}. This practice of design-by-intuition raises critical concerns, as invalidated architectural choices may inadvertently exacerbate the modality alignment discrepancy.

In this work, we take the persistent performance gap between end-to-end LSLMs and traditional ASR+LLM pipeline systems as our starting point. We systematically reproduce and quantify this discrepancy across various LLM backbones and training strategies, and, for the first time, empirically reveal the underlying mechanisms behind the performance difference. Specifically, after speech-text alignment training, a clear and consistent performance gap exists between text and speech inputs within the same model. 

To gain insight into the modality gap, we systematically analyze the similarity between speech and text representations at both sequence and token levels, aiming to reveal the mechanisms of speech-text alignment within LSLMs. At the sequence (coarse-grained) level, we observe that as representations propagate through deeper layers of the model, their cosine similarity increases steadily, reflecting progressive directional alignment. In parallel, the Euclidean distance between modalities also increases, indicating a divergence in magnitude that likely reflects modality-specific characteristics learned by the model. At the token (fine-grained) level, we find that the model develops a spontaneous monotonic alignment pattern between speech and text tokens, indicating consistent local correspondence across modalities.

In addition, our study systematically examines the relationship between internal representation similarity and the modality gap exhibited on evaluation benchmarks. A clear linear correlation is observed at both the sequence and token levels, suggesting that the nature of internal cross-modal alignment is closely related to the performance disparity between speech and text inputs.

These observations are further examined through targeted intervention experiments, where speech token embeddings along the alignment path are modified using either angle projection or length normalization. We find that such interventions can improve performance on challenging cases from the \texttt{sd-qa} subset of VoiceBench.


Our contributions are threefold: (1) We systematically identify that the primary source of the performance gap between LSLMs and pipeline systems lies in the modality difference between speech and text inputs. (2) We analyze internal representations and find that the modality gap is closely linked to the similarity between speech and text representations at both sequence and token levels. (3) We provide the first empirical evidence that targeted interventions on speech representations can improve speech input performance on challenging cases.

By focusing on understanding and revealing the mechanisms behind modality alignment, our work offers a deeper exploration of the factors influencing LSLM performance. This approach not only addresses the current performance discrepancy but also paves the way for future advancements in integrating speech modalities into LLMs.

\section{Related Work}
\subsection{Speech-Text Alignment in LSLMs}

The alignment between speech and textual modalities is crucial for the performance of LSLMs. Recent studies have explored five distinct methodologies for this task \cite{ji2024wavchatsurveyspokendialogue-survey3}. The latent space mapping approach, exemplified by \textit{Qwen2-Audio} \cite{chu2024qwen2audio}, \textit{SALMONN} \cite{tang2023salmonn}, and \textit{VITA} \cite{fu2024vita}, uses a joint audio encoder-adapter architecture to directly project speech inputs into the LLM's latent textual space. This paradigm effectively reduces computational overhead by compressing the audio sequence length via the audio adapter module. Meanwhile, it also preserves the LLM's inherent reasoning capabilities and has demonstrated competitive performance across multiple benchmarks.

Additionally, \textit{SpeechGPT} \cite{zhang2023speechgpt} adopts modality chaining by discretizing speech into symbolic units and expanding the LLM's vocabulary, while \textit{GLM-4-Voice} \cite{zeng2024glm} and \textit{Moshi} \cite{defossez2024moshi} utilize interleaved text-speech tokens and parallel generation architectures, respectively. \textit{SyncLLM} \cite{veluri2024beyond}, \textit{IntrinsicVoice} \cite{zhang2024intrinsicvoice}, \textit{Align-SLM} \cite{lin2024align}, and \textit{OmniFlatten} \cite{zhang2024omniflatten} pioneers direct speech-to-speech interaction without textual intermediates. Although significant progress has been made with these methodologies in existing research, their performance on audio processing tasks remains suboptimal.

\subsection{Modality Gap Phenomenon}
Previous work has systematically analyzed the modality gap phenomenon and shown that it persists across a wide range of multimodal models \cite{liang2022mindgapunderstandingmodality}. This gap largely arises from the cone effect, where embeddings from different modalities are restricted to distinct subspaces, leading to misalignment and degraded cross-modal performance.

In the field of speech translation, the modality gap has also been a subject of investigation. Evidence suggests that this gap can emerge during the early phases of fine-tuning \cite{han2023modality}. Furthermore, it has been shown that the resulting misalignment between modalities leads to divergent predictions and degrades performance relative to text-only machine translation systems. This representational divergence was empirically quantified using the cosine similarity between speech and text embeddings, confirming a substantial gap \cite{fang2023understandingbridgingmodalitygap}.

\subsection{Analysis of Multimodal Representations}

Previous works have explored a variety of methods to accurately quantify the similarity between multimodal representations. Canonical Correlation Analysis (CCA) and its deep learning variants, such as SVCCA and PWCCA \cite{raghu2017svcca, morcos2018insights}, are widely used to capture linear and non-linear relationships in cross-modal representations.

To address the challenges of comparing high-dimensional representations, Centered Kernel Alignment (CKA) was introduced as a robust similarity measure \cite{kornblith2019similarity}. Subsequent work has shown that a simple, sample-wise cosine similarity can also effectively capture layer-wise similarity in transformer models, yielding results comparable to CKA with greater computational efficiency \cite{jiang2024tracing}.

Additionally, the Wasserstein distance between paired speech and text embeddings has been used to measure cross-modal consistency \cite{liu2025adaptiveinnerspeechtextalignment}. The Gramian Representation Alignment Measure (GRAM) is also designed to evaluate the alignment of multiple modalities simultaneously \cite{cicchetti2024gramian}. 
Both methods have been integrated into training and effectively improve cross-modal alignment.

\section{Preliminary}
This section investigates the performance degradation of LSLMs in processing speech inputs compared to their base models' performance on text inputs. Through comprehensive experiments conducted on multiple LLM backbones using both full-parameter and LoRA fine-tuning methods \cite{hu2021loralowrankadaptationlarge}, we find that the primary contributor to this modality gap is the suboptimal alignment between textual and auditory modalities in LSLMs.

\subsection{Model Architecture}
\label{model arch}
\begin{figure}[t]
  \centering
    \includegraphics[width=\linewidth]{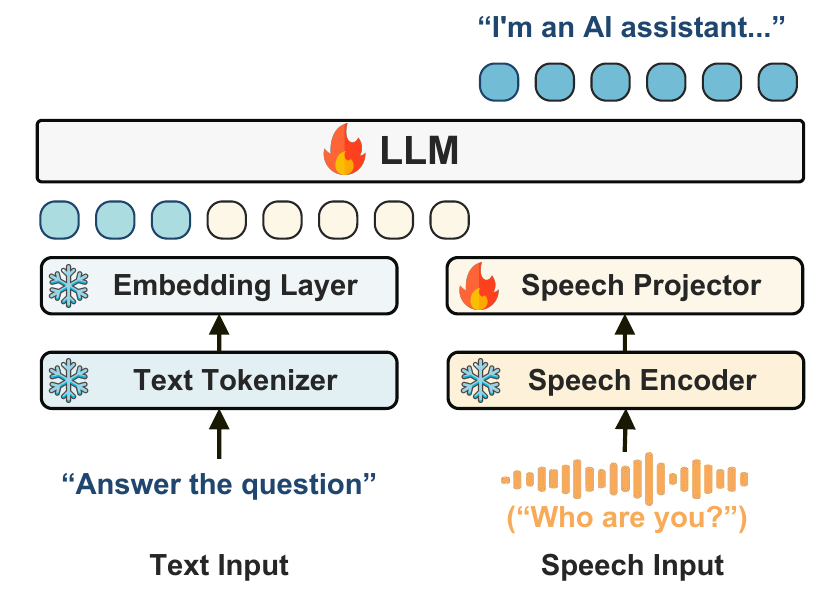}
  \caption{The LSLM architecture used in this study to analyze the speech–text modality gap.}
    \label{fig1: Model Arch}
\end{figure}

\begin{figure*}[t]
  \centering
    \includegraphics[width=1\linewidth,height=5.5cm,keepaspectratio]{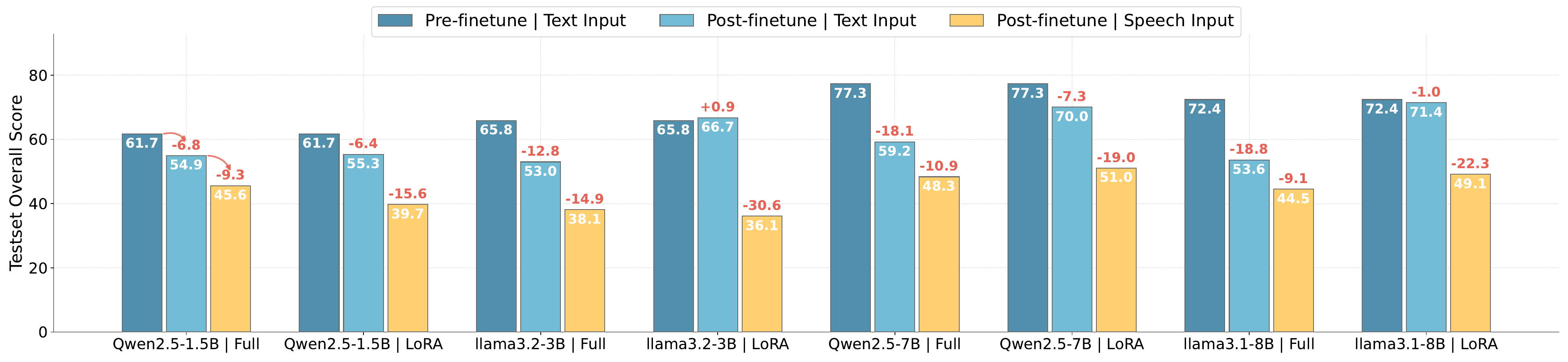}
  \caption{Overall Score (\%, $\uparrow$) on the test set of 8 LSLMs at epoch 2 under different training methods and modality inputs. The red numbers on each bar show the score difference relative to the bar immediately to its left.}
    \label{fig2: overall score bar}
\end{figure*}

As illustrated in Figure \ref{fig1: Model Arch}, the architecture of the LSLM setup used in this study comprises three core components: a speech encoder, a speech adapter, and an LLM backbone. 
The speech signal is first encoded by the speech encoder into a latent representation and then compressed by the speech adapter by a factor of \textit{5} to reduce computational overhead. 
Meanwhile, text inputs are processed through a standard tokenization pipeline and embedded via the LLM's embedding layer. 
These speech and text embeddings are then concatenated to form a unified multimodal sequence that serves as input to the LLM backbone, enabling autoregressive generation of textual responses. 

In our experiments, \textit{Whisper-large-v3}, a widely used ASR model, serves as the speech encoder. The speech adapter is implemented as a lightweight module with two fully connected layers. We conducted experiments with various LLM backbones, including \textit{LLaMA3.2-3B-Instruct}, \textit{LLaMA3.1-8B-Instruct}, \textit{Qwen2.5-1.5B-Instruct}, and \textit{Qwen2.5-7B-Instruct}. For brevity, henceforth we omit the suffix "Instruct" when referring to these model variants.

\subsection{Experiment Setups}

Our training dataset is constructed following the framework of \textit{Ke-Speech-Chat} \cite{zhao2024advancing}, exclusively comprising single-turn dialogue samples. Each sample includes both speech and text instructions, as well as a text response. We refined the raw text using \textit{Qwen2.5-72B-Instruct} \cite{qwen2, qwen2.5}, aligning it with natural conversational patterns observed in real-world scenarios. Subsequently, a three-stage filtering mechanism was applied to purify the data, targeting safety, semantic clarity, and linguistic naturalness. The speech instruction-response pairs were synthesized using \textit{CosyVoice} \cite{cosyvoice}. Based on automated transcription via \textit{Whisper-large-v3} \cite{radford2022whisper}, speech samples exceeding a Word Error Rate (WER) threshold of 0.1 are discarded. Finally, Our training dataset contains 637,283 samples, with speech instructions totaling 1,604 hours.

All LSLMs are trained for 2 epochs on our training dataset using the AdamW optimizer with a peak learning rate of 2e-5. For LoRA, we set $r=8$, $\alpha=4.0$, and the dropout rate to $0.1$. All experiments were conducted on a distributed setup with 2 nodes, each equipped with 8 NVIDIA A100 GPUs. Training a single model requires approximately 384 GPU hours on this setup.

For evaluation, we adopt five subsets of the \textit{VoiceBench} dataset \cite{chen2024voicebench}—\texttt{AdvBench}, \texttt{IFEval}, \texttt{OBQA}, \texttt{MMSU}, and \texttt{sd-qa}—yielding a total of 4,947 test samples after filtering.
These subsets collectively cover 93\% of the full VoiceBench and are particularly suitable for robust evaluation as their metrics do not require additional LLMs, thereby minimizing variability.
All evaluations strictly adhere to the official VoiceBench evaluation protocol to ensure consistency and reproducibility.

\subsection{Results and Analysis}
\label{bench_res_ana}

We evaluate the performance of LSLMs after full-parameter and LoRA fine-tuning on the 4 models introduced in Section \ref{model arch}, each tested under both speech and text modalities. Detailed results are provided in Appendix \ref{sec:appendix1 Training Results} and \ref{app:pipeline-baselines}. Figure \ref{fig2: overall score bar} presents the performance of each model at epoch 2 on both text and speech inputs, alongside the corresponding base model’s text-only performance.

For the data shown in Figure \ref{fig2: overall score bar}, on average, LSLMs exhibit a 25\% performance decline on speech inputs relative to their base models on text. This decline can be attributed to two factors: (1) fine-tuning–induced erosion of reasoning and generation capabilities, with an average drop of 8.79\%; and (2) suboptimal speech–text alignment, with an average drop of 16.46\%. Given the variety of model sizes and tuning methods evaluated, this trend appears general. This phenomenon indicates that the observed performance degradation stems primarily from the speech–text modality gap, and that bridging this gap is crucial to enhance LSLM speech processing.
Indeed, this modality gap is not unique to our setup, as we observe a similar trend across diverse public LSLMs in Appendix~\ref{sec:appendix_other_models}.

\section{Empirical Analysis of Coarse-grained Speech-Text Representations}
\label{sec4:coarse_representation}
In this section, we examine the dynamic relationship between text and speech modality representations at a coarse-grained sequence level using similarity measurement techniques. 
Our analysis uncovers consistent patterns across various LLM architectures and training paradigms. 
Through extensive experimentation, we observe a strong linear correlation between representation alignment and performance disparities across modalities, particularly under LoRA fine-tuning, highlighting the predictive value of embedding similarity for modality gap estimation.

\begin{figure}
    \centering
    \includegraphics[width=0.95\linewidth]{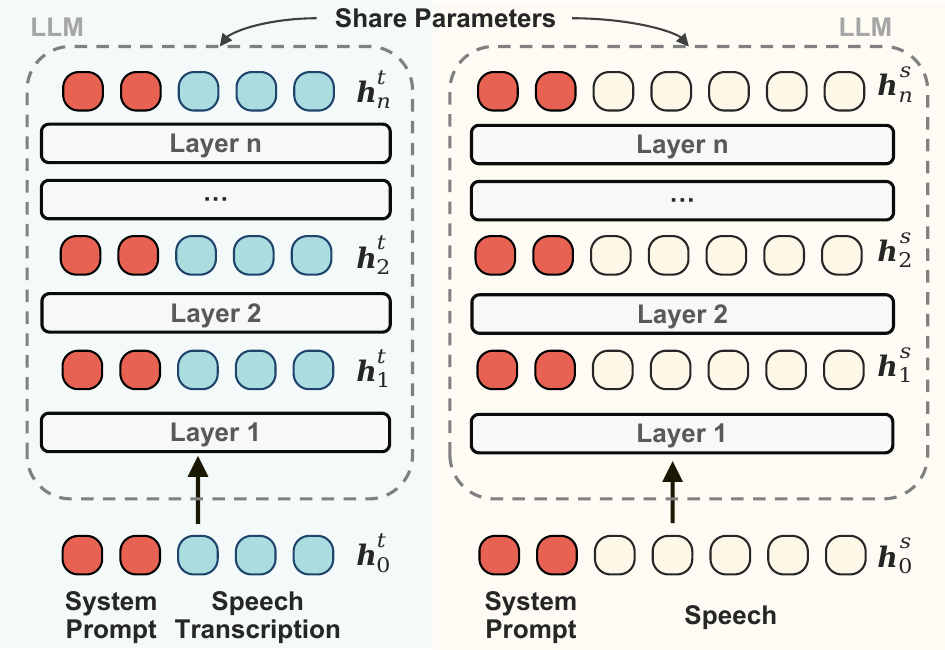}
    \caption{Overview of coarse-grained speech-text representations computation.}
    \label{fig3}
\end{figure}

\subsection{Methodology}
\label{method:avg}

Given a set of $N$ speech-text query pairs $\{(x_i^{s}, x_i^{t})\}_{i=1}^N$, where $x_i^{s}$ denotes the speech input and $x_i^{t}$ its corresponding text transcription, we process each sample through the model as illustrated in Figure~\ref{fig3}.

For the speech modality, the input $x^{s}$ is encoded by the speech encoder and linear projector, resulting in an initial embedding sequence
\(
h_0^{s} \in \mathbb{R}^{T_s \times d},
\)
where $T_s$ is the number of speech frames and $d$ is the hidden dimension. This sequence, along with a system prompt, is fed into an $L$-layer model, yielding layer-wise representations
\(
h_l^{s} \in \mathbb{R}^{T_s \times d},
\)
where $l \in \{1, \ldots, L\}$ indexes the model layer. Similarly, for the text modality, the input $x^{t}$ is tokenized and embedded, producing
\(
h_0^{t} \in \mathbb{R}^{T_t \times d},
\)
where $T_t$ is the length of the token sequence. The corresponding layer-wise representations are
\(
h_l^{t} \in \mathbb{R}^{T_t \times d}
\).

To quantify the relationship between speech and text representations, we employ two similarity metrics, denoted in a unified manner as $f^{(\cdot)}(x, y)$, where $(\cdot)$ indicates the choice of metric ($\mathrm{cos}$: cosine similarity, $\mathrm{d}$: Euclidean distance):
\[
f^{(\mathrm{cos})}(x, y) = \frac{x^\top y}{\|x\| \|y\|}, \qquad
f^{(\mathrm{d})}(x, y) = \|x - y\|_2.
\]

For each layer $l$, we first compute the mean representation over the sequence dimension for both modalities:
\[
\bar{h}_l^{s} = \frac{1}{T_s} \sum_{i=1}^{T_s} h_{l,i}^{s}, \qquad 
\bar{h}_l^{t} = \frac{1}{T_t} \sum_{j=1}^{T_t} h_{l,j}^{t},
\]
where $h_{l,i}^{s}$ and $h_{l,j}^{t}$ denote the $i$-th and $j$-th frame or token embedding at layer $l$ for the speech and text modalities, respectively. The global relationship between modalities at each layer is then assessed by computing $f_l^{(\mathrm{cos})}\left(\bar{h}_l^{s}, \bar{h}_l^{t}\right)$ and $f_l^{(\mathrm{d})}\left(\bar{h}_l^{s}, \bar{h}_l^{t}\right)$. 

\subsection{Sequence-level Speech-Text Representation Dynamics}

\begin{figure*}[t]
  \centering
  \includegraphics[width=\linewidth]{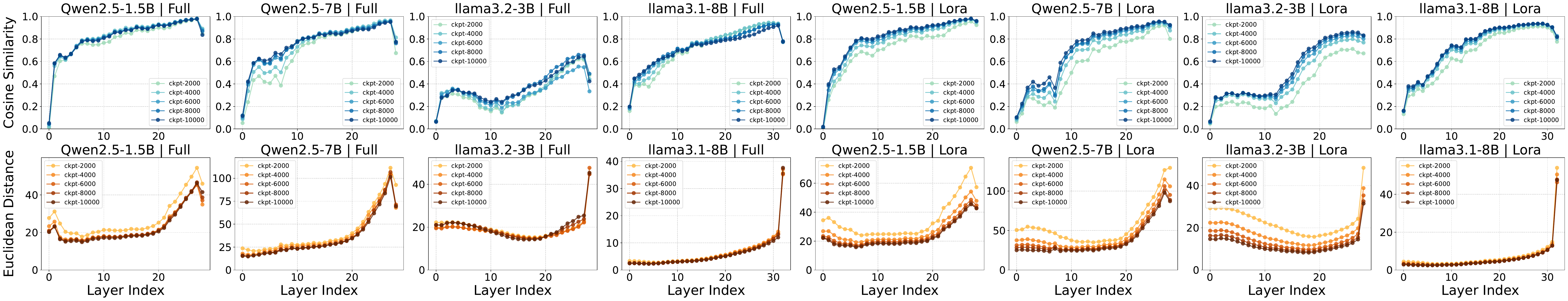}
  \caption{Layer-wise cosine similarity (blue) and Euclidean distance (orange) between speech and text representations. Each curve corresponds to a training checkpoint, with the horizontal axis indicating the model layers.}
   \label{fig2:total}
\end{figure*}
\begin{figure*}[ht]
    \centering
    \includegraphics[width=0.95\linewidth]{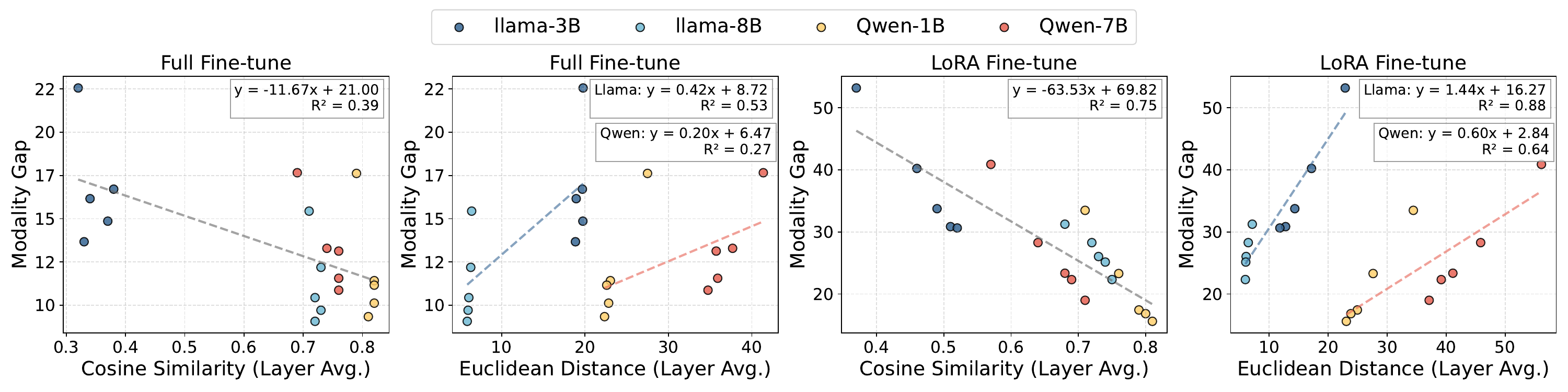}
    \caption{Linear relationship between cross-modal similarity and modality performance gap ($\textcolor{beikeblue}{GAP}$).}
    \label{fig:similarity_gap}
\end{figure*}

We evaluate the sequence-level similarity at each layer for all test samples using the methodology outlined in Section \ref{method:avg}. Figure \ref{fig2:total} summarizes these results for various LSLMs under different training regimes, with cosine similarity shown in blue and Euclidean distance in orange. Each subplot corresponds to a specific model configuration, and each curve within a subplot represents a distinct training checkpoint, depicting the layer-wise similarity metrics.

Several notable patterns are observed in the similarity dynamics.
For \textbf{cosine similarity}, all models demonstrate a consistent increase as the network depth grows, indicating progressively stronger alignment between speech and text representations in deeper layers. 
Moreover, later training checkpoints consistently yield higher similarity scores across all layers, reflecting improved cross-modal alignment as training advances.

In contrast, \textbf{Euclidean distance} gradually increases in the shallow layers and accelerates in the deeper layers. While models with more extensive training tend to exhibit slightly lower Euclidean distances overall, the upward trend with increasing depth remains consistent, suggesting growing representational magnitude divergence despite directional convergence.

These trends suggest that deeper layers and extended training foster improved alignment in representational direction (cosine similarity), while preserving modality-specific distinctions in magnitude (Euclidean distance). 
This alignment pattern could facilitate effective multimodal integration while preserving essential characteristics of each modality.

\subsection{Correlation Between Representation Similarity and Modality Gap}
\label{sec:linear_mean}
To analyze the relationship between representation similarity and downstream performance, we compute a scalar similarity score for each model by averaging similarity across layers:
\[
\bar{f}^{(\mathrm{\cdot})} = \frac{1}{L} \sum_{l=1}^{L} f_{l}^{(\mathrm{\cdot})}\left(\bar{h}_l^{s}, \bar{h}_l^{t}\right),
\]
where $(\cdot)$ denotes either cosine similarity or Euclidean distance.

We quantify the modality gap as the drop in benchmark scores between text and speech inputs, as:
\[
\textcolor{beikeblue}{GAP} = M^{t} - M^{s}, 
\]
where $M^{t}$ and $M^{s}$ are overall benchmark scores obtained from text and speech inputs, respectively.

Figure~\ref{fig:similarity_gap} shows the linear relationship between similarity and
$\textcolor{beikeblue}{GAP}$. 
Each point corresponds to a model checkpoint, with the $R^2$ value indicating the strength of correlation.

\paragraph{Key Findings.}
Under LoRA fine-tuning, a strong linear relationship is observed between cosine similarity and $\textcolor{beikeblue}{GAP}$ ($R^2 = 0.75$), suggesting that better cross-modal alignment leads to smaller performance disparities. 
Although Euclidean distance shows a weaker correlation overall, it becomes more pronounced within specific model families ($R^2 = 0.64$ for \textit{Qwen} and $R^2 = 0.88$ for \textit{Llama}).

For full-parameter fine-tuning, the same trend persists but with reduced strength: $R^2 = 0.39$ for cosine similarity, and $R^2 = 0.53$ and $0.27$ for Euclidean distance in \textit{Llama} and \textit{Qwen}, respectively.

\paragraph{Analysis.}

These findings empirically validate the connection between internal cross-modal representations and performance-level modality gaps. The stronger correlations observed in LoRA-tuned models may stem from the constrained low-rank adaptation, which preserves the integrity of pretrained text representations while facilitating targeted speech-text alignment. In contrast, full fine-tuning grants more representational flexibility, potentially introducing overfitting that weakens this correlation.
Consequently, representation similarity serves as a more reliable predictor of modality performance under LoRA than under full-parameter fine-tuning.

\section{Empirical Analysis of Finer-grained Speech-Text Representations}
Building on the coarse-grained sequence-level analysis in the previous section, this section focuses on token-level alignment patterns, examining \textbf{the role and contribution of each token in modality alignment}. We will begin with case studies, and subsequently introduce more detailed quantitative metrics to facilitate a finer-grained investigation. Through correlation analysis and intervention experiments, we explore the relationship between token-level alignment and downstream task performance, thereby further revealing the underlying speech-text alignment mechanisms in LSLMs.

\subsection{Monotonic Patterns in Token-wise Similarity Matrices}
\label{mono pattern}
\begin{figure}[t]
  \centering
  \includegraphics[width=0.48\textwidth]{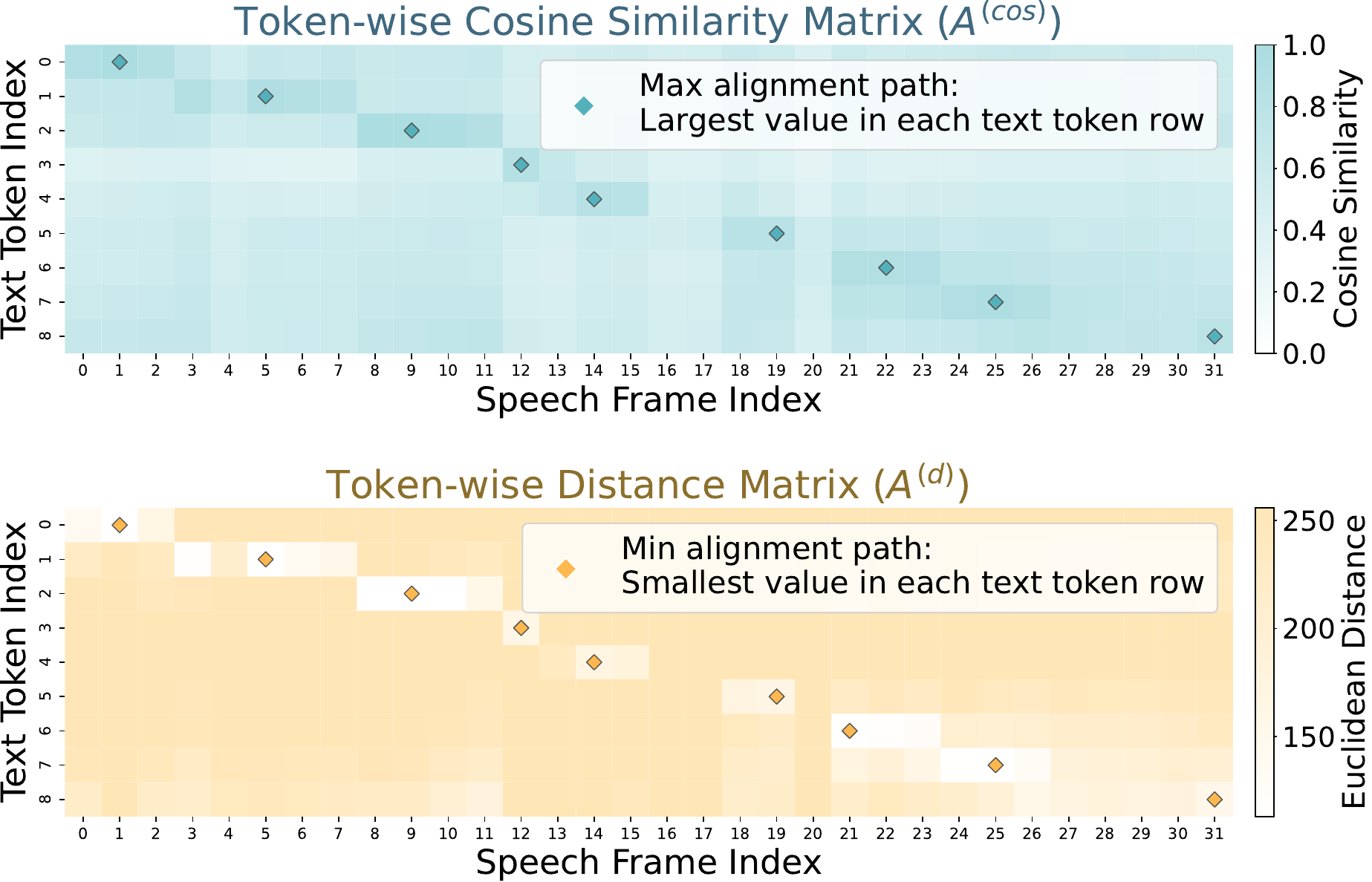}
  \caption{
    Token-wise similarity and distance matrices for a representative layer, with diamond markers indicating the alignment path.}
  \label{fig:token_alignment}
\end{figure}

\paragraph{Observations}
At each layer $l$, we construct a token-wise similarity matrix $A_l^{(\cdot)} \in \mathbb{R}^{T_s \times T_t}$, defined as: 
\[
[A_l^{(\cdot)}]_{i,j} = f^{(\cdot)}(h_{l,i}^{s}, h_{l,j}^{t}),
\]
where $f^{(\cdot)}$ denotes the selected similarity metric.
Across all models and training paradigms, we consistently observed that \textbf{the token-wise similarity matrix typically exhibits extreme values along a nearly monotonic path}. As shown in Fig.~\ref{fig:token_alignment}, with the increase in text token index, there is a monotonic alignment path in the speech frame sequence along which the similarity (or distance) values are locally maximized (or minimized). This monotonic path does not strictly align with the main diagonal, but reflects the actual temporal alignment structure between speech and text modalities.

\paragraph{Statistical Quantification}
To systematically quantify this alignment pattern, for each text token $j$, we identify the index of the speech frame with maximal similarity or minimal distance as:
\[
i^*_j = 
\begin{cases}
    \arg\max_{i} [A_l^{(\mathrm{cos})}]_{i, j} & \text{if } f^{(\cdot)} = f^{(\mathrm{cos})} \\
    \arg\min_{i} [A_l^{(\mathrm{d})}]_{i, j}   & \text{if } f^{(\cdot)} = f^{(\mathrm{d})}
\end{cases}
\]

This process produces an alignment path between the text and speech sequences. To verify the presence of monotonicity in these alignments, we use the Spearman rank correlation coefficient between text token indices and their aligned speech frame indices as the evaluation metric. Detailed statistics are provided in Appendix~\ref{app:mono}. At the final training epoch across all models, the average Spearman coefficient is 0.85 for cosine similarity and 0.70 for Euclidean distance. The proportion of tokens with perfectly identical alignment paths under both similarity measures is 0.59, indicating substantial consistency in the alignment results.

\paragraph{Mechanism Analysis}
The widespread emergence of this monotonic pattern suggests that the model not only aligns modalities globally, but also \textbf{spontaneously learns a soft, monotonic alignment between speech frames and text tokens at the token level}. Importantly, this alignment pattern emerges automatically in end-to-end speech-text alignment tasks, reflecting the model's ability to capture and map the temporal structure of speech to the semantic structure of text in a robust manner.

\subsection{Alignment Path Score}

\begin{figure*}[t]
    \centering
    \includegraphics[width=\linewidth]{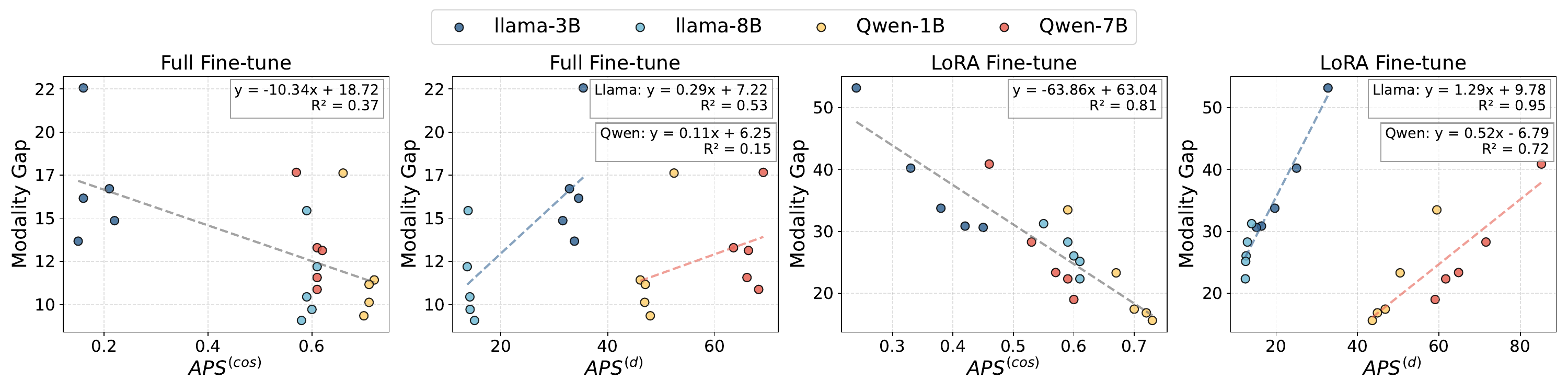}
    \caption{Relationship between the proposed Alignment Path Score and modality performance gap ($\textcolor{beikeblue}{GAP}$).}
    \label{fig:aps-regression}
\end{figure*}

\begin{table*}[t]
    \centering
    \caption{
        Performance (\%, $\uparrow$) on the \texttt{sd-qa} subset under different token-level intervention strategies.
        ``\textit{Bottom3}'': Only the three least-aligned tokens are modified; ``\textit{All}'': The entire alignment path is modified.
        For each row, results that outperform the corresponding \textit{Speech Input} are typeset in bold.
    }
    \label{tab:intervention_results}
    \resizebox{0.9\textwidth}{!}{
    \begin{tabular}{cccccccc}
        \toprule
        \multirow{2}{*}{\textbf{Model}} &
        \multirow{2}{*}{\textbf{Strategy}} &
        \multirow{2}{*}{\textbf{Text Input}} &
        \multirow{2}{*}{\textbf{Speech Input}} &
        \multicolumn{2}{c}{\textbf{Angle Projection}} &
        \multicolumn{2}{c}{\textbf{Length Normalization}} \\
        \cmidrule(lr){5-6} \cmidrule(lr){7-8}
        & & & & \textit{Bottom3} & \textit{All} & \textit{Bottom3} & \textit{All} \\
        \midrule
        \multirow{2}{*}{\textit{Qwen2.5-7B}} & Full & 43.58 & 33.45 & \textbf{38.52} & \textbf{38.70} & \textbf{37.25} & 32.73 \\
                                             & LoRA & 47.92 & 38.88 & \textbf{40.51} & 38.88 & 38.16 & 31.10 \\
        \midrule
        \multirow{2}{*}{\textit{Llama3.1-8B}} & Full & 41.05 & 36.53 & 32.91 & 34.00 & 31.65 & 30.74 \\
                                              & LoRA & 53.16 & 45.75 & \textbf{49.19} & \textbf{47.74} & 42.86 & 39.96 \\
        \bottomrule
    \end{tabular}
    }
\end{table*}

Based on the observed token-level alignment patterns, we propose the \textbf{Alignment Path Score} (APS) to quantify the strength of speech-text alignment at the token level. Specifically, APS is defined as:
\[
\mathrm{APS}^{(\cdot)} = \frac{1}{L\times T_t} \sum_{l=1}^{L} \sum_{j=1}^{T_t} [A_l^{(\cdot)}]_{i^*_j, j},
\]
where $L$ denotes the number of layers, $T_t$ is the number of text tokens, and $[A_l^{(\cdot)}]_{i^*_j, j}$ represents the maximal similarity (or minimal distance) along the alignment path for each token.

We systematically evaluate the relationship between APS and $\textcolor{beikeblue}{GAP}$ defined in Section \ref{sec:linear_mean} on the LSLMs using a linear regression analysis. As shown in Figure~\ref{fig:aps-regression}, LoRA-trained LSLMs yield higher $R^2$ values for both cosine ($0.81$ vs. $0.75$) and Euclidean APS ($0.72$ vs. $0.64$ for \textit{Qwen}; $0.95$ vs. $0.88$ for \textit{Llama}) compared to previous baselines, indicating stronger linear correlations with $\textcolor{beikeblue}{GAP}$. Under full-parameter finetuning, the correlation between APS and $\textcolor{beikeblue}{GAP}$ is similar to that of sequence-level metrics, with both showing low $R^2$ values. As previously suggested, the greater training noise and instability in this setting may limit the explanatory power of both sequence-level and token-level alignment metrics.

These results suggest that APS offers a more direct and sensitive measurement of the relationship between alignment quality and downstream performance. The stronger correlation between APS and $\textcolor{beikeblue}{GAP}$ highlights that fine-grained, token-level alignment is the key mechanism underlying LSLM speech understanding.

\subsection{Intervention Experiments: Probing the Causal Role of Token-level Alignment}
\label{interventionexp}
As demonstrated in the previous section, our analyses revealed a strong correlation between the token-level alignment score (APS) and the modality gap in model performance ($\textcolor{beikeblue}{GAP}$). However, correlation does not necessarily imply causation. To further investigate whether the token-level alignment mechanism causally affects the speech understanding ability of LSLMs, we conducted a series of targeted intervention experiments.

Specifically, we focused on the \texttt{sd-qa} subset of VoiceBench and selected both \textit{Qwen2.5-7B} and \textit{Llama3.1-8B} models, each under LoRA and full-parameter fine-tuning settings. For each sample, we first used the APS path to identify the three speech tokens with the lowest alignment scores (\textit{bottom3}) as well as all tokens along the alignment path (\textit{All}). We then applied two types of interventions: (1) \textbf{Angle projection}, where the selected speech token embeddings were projected to have the same direction as their corresponding text token embeddings; and (2) \textbf{Length normalization}, where the norm of the speech token embeddings was scaled to match that of the corresponding text tokens. We evaluated the downstream QA accuracy before and after intervention.

As shown in Table~\ref{tab:intervention_results}, Angle Projection yields improvements or maintains performance in 6 out of 8 intervention settings, demonstrating that increasing the angular similarity of token-level text and speech representations can enhance downstream outcomes. For LoRA fine-tuned models, applying angle projection to either the \textit{Bottom3} or \textit{All} alignment-path tokens consistently improves results. Notably, intervening on only the \textit{Bottom3} tokens leads to more robust gains, with \textit{Llama3.1-8B} improving by 7.52\% and \textit{Qwen2.5-7B} by 4.19\%. In contrast, length normalization provides improvement in only one case, with performance declining in the remaining settings, indicating an overall detrimental effect on LSLM’s speech sequence modeling.

Further case analysis shows that angular or length-based interventions on speech tokens can correct cases where the model fails on speech input but succeeds on the corresponding text. These corrections fall into two categories: (1) resolving semantic misunderstandings from misinterpreting spoken content, and (2) rectifying factual errors despite correct semantic parsing. Representative examples for both are provided in Appendix~\ref{sec:appendix-case-analysis}, highlighting the potential of token-level interventions to improve linguistic comprehension and factual consistency for spoken queries.

\section{Conclusion}
This work systematically investigates the modality gap in LSLMs, defined as the performance disparity between speech and text inputs within the same trained model. To uncover the mechanisms behind this gap, we analyze speech-text alignment at both sequence and token levels. 
Sequence-level analysis tracks representation similarity across layers and training, establishing its linear relationship with the modality gap.
At the token level, we reveal word-frame alignment structures and propose the Alignment Path Score, which shows a stronger correlation with the proposed modality gap. Targeted intervention experiments further demonstrate that improving token-level alignment can enhance speech inference accuracy.
This study deepens understanding of how large language models process and comprehend spoken language.

\section*{Limitations}

\noindent\textbf{Generalizability of Findings.}
Our main experiments focus on a specific set of architectures and alignment frameworks. Although Appendix~\ref{sec:appendix_B} reports consistent phenomena across external LSLMs, broader validation on larger models and fundamentally novel alignment strategies remains necessary.

\noindent\textbf{Scope of Data and Tasks.}
Our evaluation centers on English, single-turn dialogue with synthetic speech, which may limit the applicability of the findings to other languages, multi-turn conversations, and noisy real-world inputs.

\noindent\textbf{Post-hoc Nature of Interventions.}
Our intervention strategies are applied post hoc at inference time and serve primarily as analytical probes of token-level alignment. An important direction is to integrate these insights into training to explicitly optimize cross-modal consistency and improve speech-input performance.

\bibliography{custom}

\clearpage
\appendix
\section{Primary Experimental Details}
\subsection{Speech vs. Text Performance Across Training Strategies}
\label{sec:appendix1 Training Results}

Table~\ref{tab:alignment_results} presents the evaluation results of our models under different training paradigms and checkpoints. For each model and training strategy, we report the performance on both speech input and text input across multiple benchmark subsets, as well as their respective overall scores. Additionally, we provide the $\textcolor{beikeblue}{GAP}$ metric, defined as the difference between the overall text input and speech input performance. This comprehensive comparison allows us to assess the alignment and robustness of various models and training approaches with respect to both input modalities.

\begin{table*}[t!]
\centering
\small
\resizebox{\textwidth}{!}{
\begin{tabular}{ccccccccccccccccc}
\toprule
\multirow{2}{*}{Model} & \multirow{2}{*}{Strategy}  & \multirow{2}{*}{Param} & \multirow{2}{*}{Steps} 
& \multicolumn{6}{c}{Speech Input (\%, $\uparrow$)} 
& \multicolumn{6}{c}{Text Input (\%, $\uparrow$) } 
& \multirow{2}{*}{$\textcolor{beikeblue}{GAP}$ ($\downarrow$)}\\
\cmidrule(lr){5-10} \cmidrule(lr){11-16}
& & &  & AdvBench & IfEval & OBQA & MMSU & sd-qa & Overall 
& AdvBench & IfEval & OBQA & MMSU & sd-qa & Overall & \\
\midrule
\midrule
\multirow{10}{*}{Qwen2.5-1.5B}  
  & \multirow{5}{*}{Full} & \multirow{5}{*}{1.5B} 
  & 2,000  & 77.50  & 15.55 & 25.27 & 26.25 & 31.83 & 35.28 & 96.73 & 20.16 & 61.10  & 42.06 & 44.44 & 52.90 & 17.62 \\
  & & & 4,000  & 95.77 & 14.01 & 44.40 & 29.99 & 31.10 & 43.06 & 98.85 & 20.64 & 69.01 & 44.99 & 38.89 & 54.48 & 11.42 \\
  & & & 6,000  & 94.04 & 12.39 & 45.93 & 32.60 & 29.84 & 42.96 & 97.69 & 18.97 & 69.45 & 45.02 & 39.47 & 54.12 & 11.16 \\
  & & & 8,000  & 95.77 & 13.67 & 43.52 & 31.36 & 30.02 & 42.87 & 98.65 & 18.44 & 68.13 & 45.25 & 34.45 & 52.99 & 10.12 \\
  & & & 10,000 & 98.46 & 13.36 & 49.45 & 33.12 & 33.45 & 45.57 & 99.04 & 18.64 & 69.89 & 46.39 & 40.60 & 54.91 & 9.34 \\
  \cmidrule(lr){2-17}
  & \multirow{5}{*}{LoRA} & \multirow{5}{*}{9M} 
    & 2,000  & 32.17 & 11.40 & 25.71 & 26.61 & 23.33 & 23.84 & 88.08 & 32.16 & 73.85 & 52.60 & 39.96 & 57.33 & 33.48 \\
  & & & 4,000  & 59.35 & 13.67 & 30.99 & 26.28 & 27.85 & 31.63 & 84.42 & 24.27 & 74.51 & 74.51 & 39.42 & 54.93 & 23.30 \\
  & & & 6,000  & 80.00 & 14.22 & 36.04 & 27.00 & 29.48 & 37.35 & 88.08 & 21.90 & 74.07 & 51.89 & 37.97 & 54.78 & 17.43 \\
  & & & 8,000  & 83.08 & 13.81 & 36.48 & 27.94 & 30.20 & 38.30 & 88.46 & 21.82 & 74.95 & 51.76 & 38.70 & 55.14 & 16.83 \\
  & & & 10,000 & 84.42 & 13.46 & 40.00 & 30.32 & 30.38 & 39.72 & 88.08 & 21.56 & 75.38 & 51.43 & 40.14 & 55.32 & 15.60 \\
\midrule
\multirow{10}{*}{Qwen2.5-7B}
  & \multirow{5}{*}{Full} & \multirow{5}{*}{7B} 
    & 2,000  & 92.31 & 14.85 & 43.30 & 31.10 & 36.17 & 43.54 & 98.27 & 23.17 & 80.00 & 60.44 & 44.12 & 61.20 & 17.66 \\
  & & & 4,000  & 96.54 & 16.14 & 54.29 & 33.96 & 33.82 & 46.95 & 99.62 & 23.29 & 78.68 & 58.36 & 41.23 & 60.23 & 13.29 \\
  & & & 6,000  & 94.23 & 18.60 & 57.80 & 36.92 & 36.35 & 48.78 & 99.23 & 26.55 & 79.56 & 59.17 & 45.03 & 61.91 & 13.13 \\
  & & & 8,000  & 97.31 & 16.58 & 58.68 & 35.88 & 32.19 & 48.13 & 99.04 & 23.80 & 75.16 & 58.65 & 41.77 & 59.69 & 11.56 \\
  & & & 10,000 & 98.85 & 15.84 & 58.02 & 35.56 & 33.45 & 48.34 & 100.0 & 24.39 & 71.87 & 56.25 & 43.58 & 59.22 & 10.87 \\
  \cmidrule(lr){2-17}
  & \multirow{5}{*}{LoRA} & \multirow{5}{*}{20M} 
    & 2,000  & 67.12 & 14.16 & 27.91 & 26.28 & 35.44 & 34.18 & 99.04 & 62.30 & 87.47 & 69.06 & 57.50 & 75.08 & 40.89 \\
  & & & 4,000  & 92.12 & 16.92 & 42.64 & 30.03 & 35.62 & 43.47 & 99.23 & 55.38 & 88.13 & 68.41 & 47.56 & 71.74 & 28.28 \\
  & & & 6,000  & 94.81 & 22.10 & 48.13 & 34.16 & 37.07 & 47.25 & 99.04 & 50.79 & 87.25 & 67.86 & 48.10 & 70.61 & 23.35 \\
  & & & 8,000  & 95.19 & 22.63 & 49.45 & 35.56 & 38.52 & 48.27 & 99.04 & 49.71 & 88.13 & 67.79 & 48.28 & 70.59 & 22.32 \\
  & & & 10,000 & 96.92 & 25.42 & 56.26 & 37.67 & 38.88 & 51.03 & 99.04 & 48.19 & 87.69 & 67.24 & 47.92 & 70.02 & 18.99 \\
\midrule
\multirow{10}{*}{Llama3.2-3B} 
  & \multirow{5}{*}{Full} & \multirow{5}{*}{3B} 
    & 2,000  & 75.38 & 12.14 & 23.30 & 25.08 & 27.12 & 32.61 & 98.27 & 26.13 & 63.52 & 45.22 & 42.68 & 55.16 & 22.56 \\
  & & & 4,000  & 95.96 & 12.06 & 23.52 & 25.02 & 31.46 & 37.60 & 98.46 & 21.90 & 63.96 & 43.82 & 40.69 & 53.76 & 16.16 \\
  & & & 6,000  & 99.04 & 12.97 & 27.47 & 25.47 & 34.90 & 39.97 & 99.23 & 22.43 & 61.32 & 42.55 & 42.68 & 53.64 & 13.67 \\
  & & & 8,000  & 93.85 & 13.72 & 24.84 & 24.98 & 29.84 & 37.44 & 98.65 & 24.17 & 62.20 & 42.88 & 42.86 & 54.15 & 16.71 \\
  & & & 10,000 & 98.65 & 11.60 & 22.64 & 24.98 & 32.73 & 38.12 & 99.23 & 23.88 & 59.12 & 40.53 & 42.13 & 52.98 & 14.86 \\
  \cmidrule(lr){2-17}
  & \multirow{5}{*}{LoRA} & \multirow{5}{*}{12M} 
    & 2,000  & 11.15 & 14.66 & 21.32 & 25.54 & 19.89 & 18.51 & 97.50 & 63.37 & 75.60 & 56.64 & 65.46 & 71.71 & 53.20 \\
  & & & 4,000  & 58.08 & 12.18 & 22.42 & 23.19 & 35.62 & 30.30 & 98.46 & 62.64 & 76.92 & 55.50 & 59.13 & 70.53 & 40.23 \\
  & & & 6,000  & 74.42 & 12.97 & 20.88 & 24.79 & 38.70 & 34.35 & 98.08 & 59.32 & 75.60 & 54.52 & 52.98 & 68.10 & 33.75 \\
  & & & 8,000  & 84.42 & 13.52 & 19.12 & 25.37 & 38.52 & 36.19 & 98.08 & 55.39 & 77.14 & 54.39 & 50.27 & 67.05 & 30.86 \\
  & & & 10,000 & 79.66 & 13.28 & 22.64 & 24.53 & 40.33 & 36.09 & 97.69 & 53.25 & 76.92 & 54.26 & 51.54 & 66.73 & 30.64 \\
\midrule
\multirow{10}{*}{Llama3.1-8B} 
  & \multirow{5}{*}{Full} & \multirow{5}{*}{8B} 
    & 2,000  & 93.27 & 13.67 & 30.77 & 25.80 & 34.90 & 39.68 & 99.62 & 19.55 & 68.13 & 44.37 & 43.94 & 55.12 & 15.44 \\
  & & & 4,000  & 99.23 & 14.23 & 31.43 & 27.03 & 36.89 & 41.76 & 99.62 & 18.11 & 67.91 & 41.09 & 43.04 & 53.95 & 12.19 \\
  & & & 6,000  & 97.50 & 13.42 & 38.68 & 28.43 & 37.61 & 43.13 & 99.04 & 18.71 & 64.62 & 43.33 & 42.13 & 53.57 & 10.44 \\
  & & & 8,000  & 98.85 & 12.70 & 33.63 & 27.26 & 37.97 & 42.08 & 99.81 & 17.03 & 58.90 & 39.66 & 43.58 & 51.79 & 9.71 \\
  & & & 10,000 & 99.23 & 13.94 & 44.18 & 28.59 & 36.53 & 44.49 & 99.42 & 17.91 & 67.03 & 42.39 & 41.05 & 53.56 & 9.07 \\
  \cmidrule(lr){2-17}
  & \multirow{5}{*}{LoRA}  & \multirow{5}{*}{20M} 
    & 2,000  & 85.96 & 16.60 & 27.69 & 27.62 & 43.76 & 40.33 & 90.38 & 67.68 & 81.10 & 64.09 & 54.61 & 71.57 & 31.25 \\
  & & & 4,000  & 94.23 & 17.43 & 36.04 & 30.68 & 43.94 & 44.47 & 99.42 & 63.58 & 81.54 & 64.57 & 54.61 & 72.75 & 28.28 \\
  & & & 6,000  & 93.46 & 19.16 & 41.10 & 31.10 & 45.57 & 46.08 & 99.42 & 61.10 & 81.10 & 64.18 & 54.79 & 72.12 & 26.04 \\
  & & & 8,000  & 97.12 & 19.36 & 43.74 & 32.37 & 43.58 & 47.23 & 99.42 & 62.75 & 81.10 & 64.02 & 54.61 & 72.38 & 25.15 \\
  & & & 10,000 & 96.92 & 21.12 & 49.01 & 32.69 & 45.75 & 49.10 & 99.42 & 59.16 & 81.10 & 64.31 & 53.16 & 71.43 & 22.33 \\
\bottomrule
\end{tabular}
}
\caption{Comparison of Alignment Experiment Results: Speech and Text Input Performance Across Steps}
\label{tab:alignment_results}
\end{table*}

\subsection{Performance of Pipeline System Baselines}
\label{app:pipeline-baselines}

For a comprehensive comparison with the end-to-end LSLMs analyzed in this work, we report the performance of traditional ASR+LLM pipeline systems. These systems utilize \textit{Whisper-large-v3} as the ASR module, paired with the corresponding LLM backbones from our main experiments. All pipeline evaluations were conducted on the identical 4,947-sample VoiceBench test set and adhere strictly to the official protocol to ensure a fair comparison. The results are presented in Table~\ref{tab:pipeline-performance}.

\begin{table*}[htbp!]
\centering
\small
\renewcommand{\arraystretch}{1.2}
\setlength{\tabcolsep}{8pt}
\begin{tabular}{cccccc c}
\toprule
\textbf{Model} & \textbf{AdvBench} & \textbf{IfEval} & \textbf{OBQA} & \textbf{MMSU} & \textbf{sd-qa} & \textbf{Overall} \\
\midrule
\textit{Qwen2.5-1.5B} & 97.31 & 41.82 & 69.67 & 50.78 & 49.01 & 61.72 \\
\textit{Llama3.2-3B}  & 98.08 & 69.71 & 60.66 & 51.37 & 49.28 & 65.82 \\
\textit{Qwen2.5-7B}   & 98.27 & 70.58 & 84.84 & 69.03 & 63.83 & 77.31 \\
\textit{Llama3.1-8B}  & 98.46 & 71.12 & 72.09 & 62.04 & 58.23 & 72.39 \\
\bottomrule
\end{tabular}
\caption{Performance of pipeline baselines on the VoiceBench test set. Each system consists of a \textit{Whisper-large-v3} ASR frontend followed by the specified LLM backend.}
\label{tab:pipeline-performance}
\end{table*}

\subsection{Analysis of Alignment Path Monotonicity and Consistency}
\label{app:mono}
\begin{figure*}[htbp!]
    \centering
    \includegraphics[width=\textwidth]{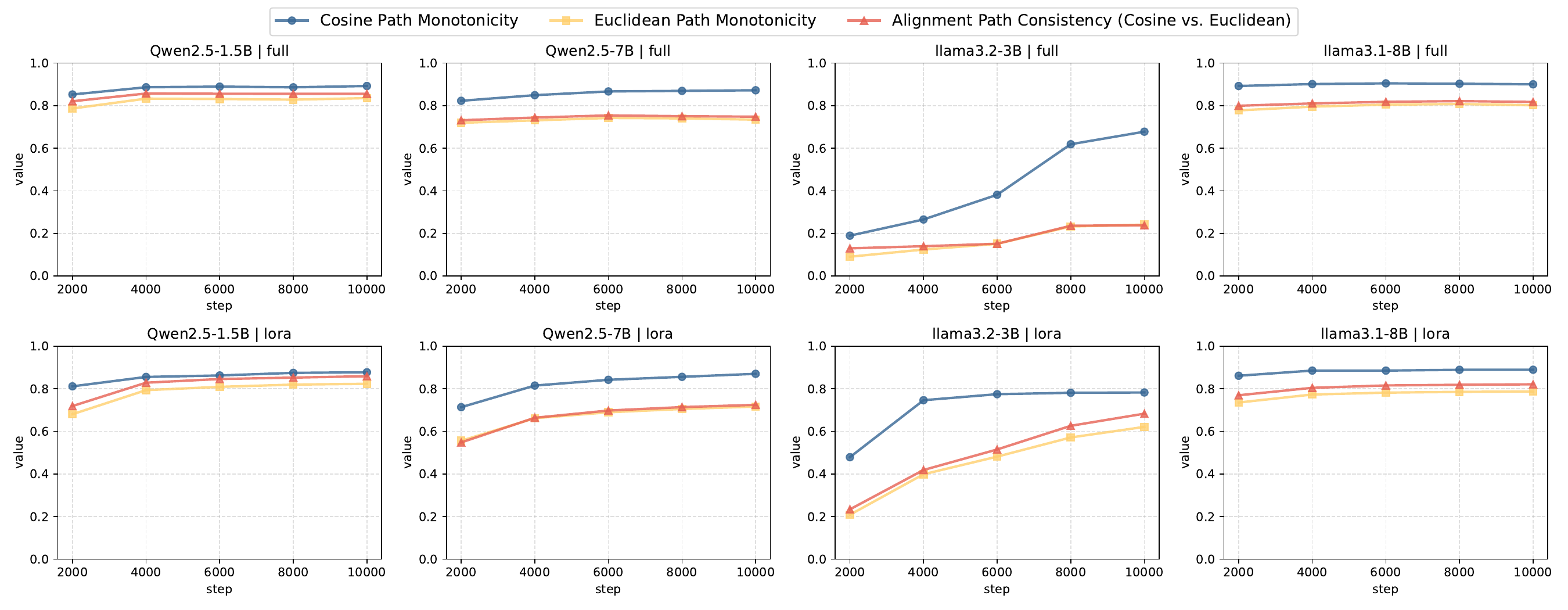}
    \caption{
        \textbf{Evolution of Alignment Path Monotonicity and Consistency.}
        For each model (\textit{columns}) and training strategy (\textit{rows}), we report the trajectory of three metrics: Cosine Path Monotonicity (blue), Euclidean Path Monotonicity (yellow), and Alignment Path Consistency (orange) across checkpoints during training. Higher monotonicity values indicate a stronger tendency toward monotonic alignments between text tokens and speech frames. Higher consistency values reflect greater agreement between alignment paths derived from the two similarity measures, suggesting stable and modality-agnostic alignment patterns.
    }
    \label{fig:mono_plot}
\end{figure*}

This section details the measurement of alignment path statistics, as introduced in Section~\ref{mono pattern}. We report these statistics across different training stages, model scales, and training strategies. As shown in Figure~\ref{fig:mono_plot}, we consider three metrics: (1) alignment path monotonicity based on the cosine similarity matrix, which reflects the degree of order in the alignment between text tokens and speech frames; (2) alignment path monotonicity based on the Euclidean distance matrix, defined in a similar manner but using Euclidean distances for alignment construction; and (3) token-level alignment path consistency, defined as the proportion of tokens whose aligned speech frame indices are identical under both similarity measures.

Empirical results reveal the following trends: (1) Both alignment path monotonicity metrics exhibit an overall increasing tendency as training progresses, suggesting that the model incrementally acquires more structured and monotonic alignments. (2) The monotonicity measured via cosine similarity remains consistently higher than that based on Euclidean distance, indicating that cosine similarity may be more effective in capturing ordered relationships in high-dimensional spaces. (3) Token-level alignment path consistency also demonstrates a general upward trend during training, implying that the alignment paths derived from the two similarity measures become increasingly similar. These observations are consistent across different model scales and training strategies, underscoring the robustness and effectiveness of the learned alignment mechanism.

\subsection{Case Analysis of Intervention Experiments}
\label{sec:appendix-case-analysis}

This section presents representative cases from the intervention experiments detailed in Section~\ref{interventionexp}, with results compiled in Tables~\ref{tab:case1} through~\ref{tab:case4}. The interventions involved two primary strategies: \textbf{angle} projection, which aligns the direction of the speech representation with its corresponding text representation, and \textbf{length} normalization, which matches their vector norms. These strategies were applied either to the three tokens with the lowest alignment confidence (\textbf{bot3}) or to all tokens along the alignment path (\textbf{all}).

Our analysis of these cases reveals two primary categories of error correction. The first category involves the resolution of semantic misunderstandings arising from the spoken input. For example, in Case 1 (Table~\ref{tab:case1}), the model initially misinterprets the spoken entity "Brittany" as "Britain," leading to an irrelevant answer. Similarly, Case 2 (Table~\ref{tab:case2}) shows an erroneous entity recognition from the speech input. After applying interventions to key tokens along the alignment path, the model successfully realigns its semantic representation with the ground-truth text, thereby recovering the correct understanding and generating an accurate response.

The second category addresses factual errors that occur even when the initial semantic parsing of the spoken query is correct. In Case 3 (Table~\ref{tab:case3}) and Case 4 (Table~\ref{tab:case4}), the model demonstrates a correct understanding of the question's topic but fails to produce factually accurate or complete answers. Our interventions, by modifying the representation direction or norm along the alignment path, also prove effective in these scenarios, guiding the model to generate factually correct responses consistent with the reference answers.

Collectively, these case studies demonstrate that fine-grained interventions on the token alignment path, through either embedding direction or norm modification, consistently improve the model's answer accuracy and robustness for spoken inputs. This effect is observed in correcting both semantic misinterpretations and factual knowledge errors, indicating that such interventions can enhance multimodal alignment and enable more reliable knowledge retrieval from speech input.

\begin{table*}[htbp]
\centering
\begin{tcolorbox}[colback=gray!2!,colframe=gray!80!black,title=Case 1]
\textbf{Question:} What do they speak in \textcolor{red}{Brittany}?

\vspace{4pt} 
\noindent\hrule 
\vspace{4pt}

\textbf{Reference Answer:} Breton.

\vspace{4pt} 
\noindent\hrule 
\vspace{4pt}

\textbf{Text Input:} In \textcolor{red}{Brittany}, the local language is Breton. It's a Celtic language that's still spoken by some people in the region. (\cmark)

\textbf{Speech Input:} In \textcolor{red}{Britain}, the primary language spoken is English. However, there are also other languages spoken, especially in areas with diverse populations, such as Welsh in Wales, Scottish Gaelic in Scotland, and Irish in Northern Ireland. (\xmark)

\textbf{Speech Input (Angle Bot3):} Breton is a Celtic language that's still spoken in parts of France, particularly in the region of \textcolor{red}{Brittany}. It has a unique sound and rhythm that's quite distinct from other languages. (\cmark)

\textbf{Speech Input (Length Bot3):} in Scotland, they speak Scottish Gaelic. In Northern Ireland, they speak Northern Irish, which is a dialect of Irish. In Wales, they speak Welsh. And in Cornwall, they speak Cornish. Each place has its own unique language and cultural heritage. (\xmark)

\textbf{Speech Input (Angle All):} In \textcolor{red}{Brittany}, they primarily speak French. However, there are some areas where you might also hear Breton, which is a Celtic language. But don't worry, most people there will be speaking French. (\cmark)

\textbf{Speech Input (Length All):} Cockney, which is a type of East London accent. It's known for its unique way of pronouncing certain letters and sounds, often making it quite distinct from other accents in the United Kingdom. (\xmark)
\end{tcolorbox}
\caption{Case 1 from intervention experiments on the \texttt{sd-qa} dataset.}
\label{tab:case1}
\end{table*}

\begin{table*}    
\begin{tcolorbox}[colback=gray!2,colframe=gray!80!black,title=Case 2]
\textbf{Question:} Who is \textcolor{red}{Sean Hannity}?

\vspace{4pt} 
\noindent\hrule 
\vspace{4pt}

\textbf{Reference Answer:} An American talk show host and conservative political commentator.

\vspace{4pt} 
\noindent\hrule 
\vspace{4pt}

\textbf{Text Input:} \textcolor{red}{Sean Hannity} is an American conservative political commentator and radio host. He's best known for hosting his own show on Fox News, where he discusses politics, current events, and social issues. He's also a popular author and has written several books. (\cmark)

\textbf{Speech Input:} \textcolor{red}{Shawn Hanity} is an American actor and comedian. He's best known for his roles in television shows like Saturday Night Live and Brooklyn Nine-Nine. He's also appeared in several films and has been involved in various other projects. (\xmark)

\textbf{Speech Input (Angle Bot3):} \textcolor{red}{Sean Hannity} is a well-known American conservative political commentator and television personality. He hosts a popular radio show and has his own talk show on Fox News, where he often discusses current political issues and provides commentary on various topics. (\cmark)

\textbf{Speech Input (Length Bot3):} \textcolor{red}{Shan}? He's a British comedian and actor known for his work in both stand-up and on television shows like The League of Gentlemen. (\xmark)

\textbf{Speech Input (Angle All):} \textcolor{red}{Sean Hannity} is a well-known American conservative political commentator and television personality. He hosts a popular radio show and has his own talk show on Fox News, where he often discusses current political issues and supports Republican candidates. (\cmark)

\textbf{Speech Input (Length All):} \textcolor{red}{Sean Hannity} is a well-known American conservative political commentator and television personality. He hosts a popular radio show and has his own talk show on Fox News, where he often discusses current political issues and provides commentary on various topics. Sean Hannity is also a best-selling author and has written several books. (\cmark)
\end{tcolorbox}
\caption{Case 2 from intervention experiments on the \texttt{sd-qa} dataset.}
\label{tab:case2}
\end{table*}

\begin{table*}    
\begin{tcolorbox}[colback=gray!2,colframe=gray!80!black,title=Case 3]
\textbf{Question:} What's the \textcolor{red}{biggest wasp}?

\vspace{4pt} 
\noindent\hrule 
\vspace{4pt}

\textbf{Reference Answer:} Asian giant hornet.

\vspace{4pt} 
\noindent\hrule 
\vspace{4pt}
\textbf{Text Input:} The \textcolor{red}{largest wasp} in the world is the Asian giant hornet. It can grow up to two and a half inches long, which is quite big compared to other wasps. (\cmark)

\textbf{Speech Input:} The \textcolor{red}{largest wasp} is the tarantula hawk wasp, which can grow up to two and a half inches long. These wasps are known for their impressive size and their ability to paralyze tarantulas to use as food for their larvae. (\xmark)

\textbf{Speech Input (Angle Bot3):} The \textcolor{red}{largest wasp} species is the giant hornet, also known as the Asian giant hornet. These impressive insects can have a wingspan of up to two inches and are found in various parts of Asia. They are known for their aggressive behavior and can be quite dangerous. (\cmark)

\textbf{Speech Input (Length Bot3):} It's called the Great Barrier Reef, and it's located off the coast of Australia. It's so big that it can even be seen from space! (\xmark)

\textbf{Speech Input (Angle All):} The \textcolor{red}{largest wasp} species is the golden paper wasp. They can build some of the largest and most complex nests among all wasp species, often using chewed wood fibers mixed with saliva to create a papery material. These nests can be quite impressive! (\xmark)

\textbf{Speech Input (Length All):} It's called the Great Barrier Reef, and it's located off the coast of Australia. It's so big that it can even be seen from space! (\xmark)
\end{tcolorbox}
\caption{Case 3 from intervention experiments on the \texttt{sd-qa} dataset.}
\label{tab:case3}
\end{table*}

\begin{table*}    
\begin{tcolorbox}[colback=gray!2,colframe=gray!80!black,title=Case 4]
\textbf{Question:} Where does the story of American Horror Story: \textcolor{red}{Murder House} take place?

\vspace{4pt} 
\noindent\hrule 
\vspace{4pt}
\textbf{Reference Answer:} Los Angeles, California.

\vspace{4pt} 
\noindent\hrule 
\vspace{4pt}
\textbf{Text Input:} The story of American Horror Story: \textcolor{red}{Murder House} takes place in Los Angeles, California. (\cmark)

\textbf{Speech Input:} The story of American horror story \textcolor{red}{murder house} takes place in the fictional town of Harmony, which is modeled after the real town of New Orleans. The house itself is located in the French Quarter, known for its rich history and eerie atmosphere. (\xmark)

\textbf{Speech Input (Angle Bot3):} American Horror Story \textcolor{red}{Murder House} is set in a fictional address in Los Angeles, California. The exact location isn't specified, but it's portrayed as being in the heart of the city. (\cmark)

\textbf{Speech Input (Length Bot3):} American Horror Story \textcolor{red}{Murder House} is set in a fictional address in Los Angeles, California. The exact location isn't specified, but it's portrayed as being in the heart of the city. (\cmark)

\textbf{Speech Input (Angle All):} American Horror Story \textcolor{red}{Murder House} is set in New York City. (\xmark)

\textbf{Speech Input (Length All):} American Horror Story \textcolor{red}{Murder House} is set in New York City. The story revolves around a haunted house in the heart of the city, where the main character, a young woman named Wendy, moves in with her husband and their son. The house has a dark history, and its past is gradually revealed throughout the series. (\xmark)
\end{tcolorbox}
\caption{Case 4 from intervention experiments on the \texttt{sd-qa} dataset.}
\label{tab:case4}
\end{table*}

\section{Validation on External LSLM}
\label{sec:appendix_B}
\subsection{Modality Gap Across Diverse Paradigms}
\label{sec:appendix_other_models}

To substantiate the generality of the modality gap, we extended our analysis to several publicly available LSLMs that represent diverse alignment paradigms. These models include \textit{SpeechGPT} \cite{zhang2023speechgpt}, which relies on speech discretization; \textit{BLSP} \cite{wang2024blspbootstrappinglanguagespeechpretraining}, which aligns the speech and text modalities via bootstrapped behavior alignment; \textit{GLM-4-Voice} \cite{zeng2024glm}, which utilizes interleaved text-speech tokens; and \textit{Qwen2-Audio} \cite{chu2024qwen2audio}, which employs a three-stage training pipeline (pre-training, SFT, and DPO) and uses natural language prompts to unify large-scale audio tasks during pre-training. Each model was evaluated on our standardized VoiceBench test set, comparing performance on speech inputs against their corresponding text transcriptions. 

The results, presented in Table \ref{tab:other_models_gap}, consistently reveal a significant performance drop for speech inputs across all models. This corroborates our central thesis that the modality gap is a prevalent challenge, independent of the specific LSLM architecture or alignment strategy.

\begin{table*}[t]
\centering
\small
\caption{Performance comparison of various LSLMs on speech and text inputs. }
\label{tab:other_models_gap}
\setlength{\tabcolsep}{8pt}
\begin{tabular}{llccccccc}
\toprule
\textbf{Model} & \textbf{Input Type} & \textbf{AdvBench} & \textbf{IfEval} & \textbf{OBQA} & \textbf{MMSU} & \textbf{sd-qa} & \textbf{Overall} & $\textcolor{beikeblue}{GAP}$ ($\downarrow$) \\
\midrule
\multirow{2}{*}{\textbf{SpeechGPT}} & Speech & 85.00 & 18.41 & 25.27 & 25.60 & 26.04 & 36.06 & \multirow{2}{*}{8.09} \\
& Text & 87.69 & 23.82 & 28.35 & 25.57 & 55.33 & 44.15 & \\
\midrule
\multirow{2}{*}{\textbf{BLSP}} & Speech & 8.65 & 18.01 & 21.10 & 24.37 & 49.73 & 24.37 & \multirow{2}{*}{15.65} \\
& Text & 7.88 & 37.00 & 52.31 & 37.61 & 65.28 & 40.02 & \\
\midrule
\multirow{2}{*}{\textbf{GLM-4-Voice}} & Speech & 79.81 & 24.93 & 51.43 & 38.58 & 53.89 & 49.73 & \multirow{2}{*}{6.51} \\
& Text & 85.38 & 31.18 & 60.00 & 42.62 & 62.03 & 56.24 & \\
\midrule
\multirow{2}{*}{\textbf{Qwen2-Audio}} & Speech & 97.69 & 19.95 & 42.42 & 35.82 & 43.58 & 47.89 & \multirow{2}{*}{13.30} \\
& Text & 98.85 & 28.60 & 70.11 & 44.57 & 63.83 & 61.19 & \\
\bottomrule
\end{tabular}
\end{table*}

\subsection{In-depth Analysis of Modality Alignment in \textit{Qwen2-Audio}}
\label{subsec:qwen2_audio_analysis}

To further validate the robustness of our analytical framework and conclusions across different training paradigms, we conducted an in-depth analysis of \textit{Qwen2-Audio} \cite{chu2024qwen2audio}. This model is particularly representative as it utilizes a multi-stage training pipeline involving pre-training, supervised fine-tuning, and direct preference optimization. As demonstrated below, despite its distinct training methodology, \textit{Qwen2-Audio} exhibits patterns in its modality alignment mechanism that are highly consistent with our core findings.

\begin{figure}[h!]
    \centering
    \includegraphics[width=1\columnwidth]{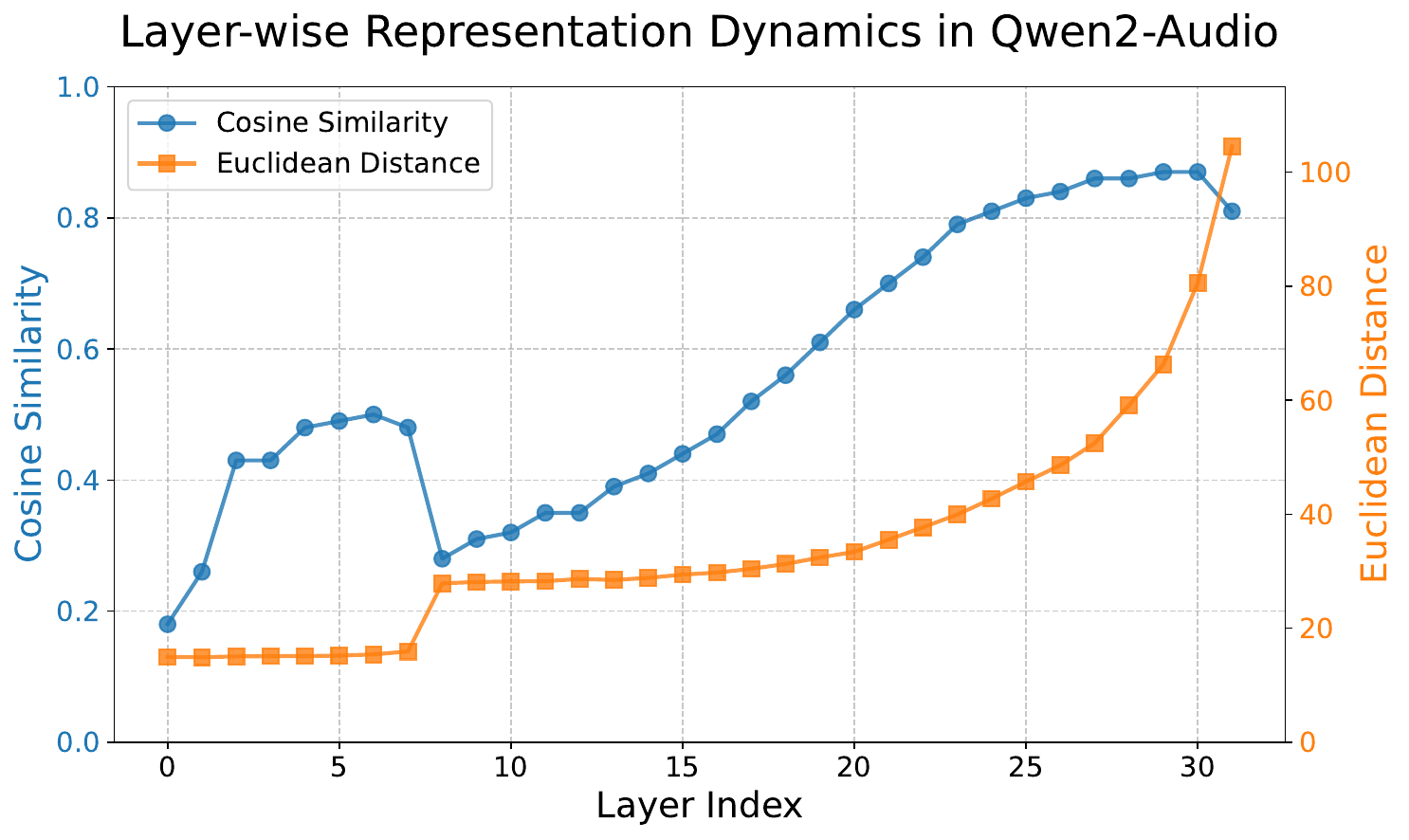}
    \caption{
        Layer-wise cosine similarity (blue) and Euclidean distance (orange) between speech and text representations for \textit{Qwen2-Audio}. The horizontal axis indicates the model layers.
    }
    \label{fig:qwen2_audio_similarity}
\end{figure}

\paragraph{Coarse-Grained Representation Dynamics.}
Following the methodology outlined in Section~\ref{sec4:coarse_representation}, we analyzed the sequence-level similarity dynamics between speech and text representations in \textit{Qwen2-Audio}. The results are visualized in Figure~\ref{fig:qwen2_audio_similarity}. The observed layer-wise similarity dynamics, characterized by an increase in cosine similarity and a concurrent upward trend in Euclidean distance with network depth, are highly analogous to the phenomena identified in our primary experiments.

\begin{table*}[ht!]
    \centering
    \small
    \caption{Alignment path statistics for \textit{Qwen2-Audio}.}
    \label{tab:qwen2_audio_monotonicity}
    \begin{tabular}{l ccc}
        \toprule
        \textbf{Statistic} & \textbf{Cosine Path Monotonicity} & \textbf{Euclidean Path Monotonicity} & \textbf{Alignment Path Consistency} \\
        \midrule
        Value & 0.7891 & 0.7586 & 0.6875 \\
        \bottomrule
    \end{tabular}
\end{table*}

\begin{table*}[h!]
    \centering
    \caption{
        Performance (\%, $\uparrow$) on the \texttt{sd-qa} subset for \textit{Qwen2-Audio} under different token-level intervention strategies. Results that outperform the corresponding \textit{Speech Input} are typeset in bold.
    }
    \small
    \label{tab:qwen2_audio_intervention}
    \begin{tabular}{lcccccc}
        \toprule
        \multirow{2}{*}{\textbf{Model}} &
        \multirow{2}{*}{\textbf{Text Input}} &
        \multirow{2}{*}{\textbf{Speech Input}} &
        \multicolumn{2}{c}{\textbf{Angle Projection}} &
        \multicolumn{2}{c}{\textbf{Length Normalization}} \\
        \cmidrule(lr){4-5} \cmidrule(lr){6-7}
        & & & \textit{Bottom3} & \textit{All} & \textit{Bottom3} & \textit{All} \\
        \midrule
        \textit{Qwen2-Audio} & 58.77 & 35.08 & \textbf{35.99} & 34.18 & 33.27 & 34.72 \\
        \bottomrule
    \end{tabular}%
\end{table*}

\paragraph{Fine-Grained Monotonicity Analysis.}

We further evaluated the token-level alignment path monotonicity for \textit{Qwen2-Audio} using the three metrics defined in Appendix~\ref{app:mono}. As presented in Table~\ref{tab:qwen2_audio_monotonicity}, these results are highly consistent with the values obtained from the models in the primary experiments at their final training stages, detailed in Figure~\ref{fig:mono_plot}. Such consistency across disparate training paradigms provides strong evidence for the spontaneous emergence of a monotonic alignment path as a generalizable phenomenon.

\paragraph{Token-Level Intervention Experiments.}
Finally, we replicated the token-level intervention experiments from Section~\ref{interventionexp} on the \texttt{sd-qa} subset. As shown in Table~\ref{tab:qwen2_audio_intervention}, applying angle projection to the least-aligned tokens (\textit{Bottom3}) successfully improved performance on speech inputs. This result demonstrates that our proposed intervention strategy for mitigating the modality gap is also effective for models trained with a multi-stage approach.

\end{document}